\documentclass[twoside,twocolumn]{article}
\usepackage{amsmath}
\usepackage{textcomp}     
\usepackage{wasysym}
\usepackage{graphicx}
\usepackage{blindtext} 
\usepackage{float}
\usepackage{hyperref}
\usepackage[sc]{mathpazo} 
\usepackage[T1]{fontenc} 
\usepackage{microtype} 
\usepackage[english]{babel} 
\usepackage{bm}
\usepackage[hmarginratio=1:1,top=32mm,columnsep=20pt]{geometry} 
\usepackage[hang, small,labelfont=bf,up,textfont=it,up]{caption} 
\usepackage{booktabs} 
\usepackage{listings}
\usepackage{lettrine} 
\usepackage[section]{placeins}
\usepackage{enumitem} 
\setlist[itemize]{noitemsep} 
\usepackage{caption}
\usepackage{abstract} 
\usepackage{needspace}
\usepackage{titlesec} 
\renewcommand\thesection{\Roman{section}} 
\renewcommand\thesubsection{\roman{subsection}} 
\titleformat{\section}[block]{\large\scshape\centering}{\thesection.}{1em}{} 
\titleformat{\subsection}[block]{\large}{\thesubsection.}{1em}{} 

\usepackage{fancyhdr} 
\usepackage{titling} 
\usepackage{hyperref} 

\pretitle{\begin{center}\huge\bfseries} 
\posttitle{\end{center}} 
\title{Syntactically Informed Text Compression with Recurrent Neural Networks} 
\author{%
\textsc{David Cox}\thanks{With sincere thanks to Dr. J. Maurice Rojas.} \\[1ex] 
\normalsize Texas A\&M University \\ 
\normalsize \href{mailto:davidcox143@tamu.edu}{davidcox143@tamu.edu} 
}
\date{July 2016} 
\renewcommand{\maketitlehookd}{%
\begin{abstract}
We present a self-contained system for constructing natural language models for use in text compression. Our system improves upon previous neural network based models by utilizing recent advances in syntactic parsing -- Google's SyntaxNet -- to augment character-level recurrent neural networks. RNNs have proven exceptional in modeling sequence data such as text, as their architecture allows for modeling of long-term contextual information. Modeling and coding are the backbone of modern compression schemes. While coding is considered a solved problem, generating effective, domain-specific models remains a critical step in the process of improving compression ratios. 
\end{abstract}
}

\begin{document}

\maketitle


\section{Introduction}
Accurate models are the key to better data compression. Compression algorithms operate in two steps -- modeling and coding. Coding is the reversible process of reassigning symbols in a sequence of data such that its length is reduced. Modern coding methods are currently close to the theoretically optimal limit of $\log_2 \frac{1}{p}$ bits per symbol. Modeling, on the other hand, is a provably unsolvable problem~\cite{mahoney}. Consider if this were not the case. Such a model would be able to accurately estimate the next symbol in a sequence of random (or compressed) data, and would be able to recursively compress its own output to zero bytes. Rather than search for this impossible universal model, efforts have focused on generating domain-specific models that exploit intrinsic structure within data. \\

A previous approach to generating natural language models by Matt Mahoney~\cite{mahoney2000fast} used a two layer, $4 \times 10^6 \times 1$ neural network as a substitute for prediction by partial matching -- a popular modeling method. The simplicity of this model allows it to process $10^4$ characters per second, compressing \textit{Alice in Wonderland} to 2.283 bpc, and comparing favorably to gzip at 3.250 bpc. While effective, this approach lacks the ability to model long-term relationships in character sequences and also does not take advantage of syntactic or semantic information. Mahoney notes that the ability to do so was one of the reasons he chose to use neural networks in the first place.  \\

We address these limitations through the use of a recurrent neural network architecture and by utilizing Google's SyntaxNet~\cite{andor2016globally} to provide part of speech annotations. Our model processes sequences of characters and part of speech tags using separate recurrent layers. The output of these layers is then merged and processed with a final recurrent layer and two fully connected layers. We found that such an architecture was able to reliably predict the next character in a sequence of forty characters without explicitly memorizing the training data when provided documents of sufficient length. While our aim was to construct probability models tailored to specific input data, our results indicate that acceptable performance can be obtained from a generalized model. A generalized natural language model would be highly desirable as it would allow for neural network based compression without the need to train a model for each input document. This publication aims to serve as the foundational work for such a model.


\section{Background}
\subsection*{A note for MAA MathFest}
This paper relies heavily on concepts from computer science. To ensure that it is accessible to our audience at MathFest, we've written it to be as self-containing as possible.  
\subsection{Data Compression}
As mentioned, data compression consists of two steps -- modeling and coding. Arithmetic coding~\cite{witten1987arithmetic} is a near-optimal coding method that operates by representing a sequence of probabilities as a fractional number in the interval $[0,1)$. \\

To illustrate arithmetic coding, consider the following example:
\begin{itemize}
    \item Let $M$, a message to be compressed, be the sequence of symbols [C,\;O,\;D,\;E,\;!].
    \item Let $S$ be an alphabet containing the symbols \{A,\;B,\;C,\;D,\;E,\;O,\;!\}.
\end{itemize}
The following table contains an arbitrary fixed probability model for the alphabet $S$:
\begin{table}[H]
\centering
\begin{tabular}{|c|c|l|}
\hline
Symbol & Probability & Range        \\ \hline
A      & 0.3         & {[}0, 0.3)    \\
B      & 0.2         & {[}0.3, 0.5) \\
C      & 0.2         & {[}0.5, 0.7) \\
D      & 0.1         & {[}0.7, 0.8) \\
E      & 0.1         & {[}0.8, 0.9) \\
O      & 0.05        & {[}0.9, 0.95) \\ 
!      & 0.05        & {[}0.95, 1.0) \\ \hline
\end{tabular}
\caption{Fixed probability model for $S$}
\end{table}

We encode $M$ by reducing the range of our subinterval from its initial range of $[0,1)$ for each symbol as shown in Table 2.

\begin{table}[H]
\centering
\begin{tabular}{|c|c|}
\hline
$\diameter$ & {[0,1)}\\ 
After C       & {[0.5, 0.7) }\\
O             & {[0.68, 0.690) }\\
D             & {[0.687, 0.688) }\\
E             & {[0.6878, 0.6879) }\\
!             & {[0.687895, 0.68790) }\\
CODE!         & {0.687895} \\\hline
\end{tabular}
\caption{Step by step arithmetic coding of $M$}
\end{table}

The final subinterval after following the steps listed above is $[0.687895, 0.68790)$. Any number within the subinterval can be decoded by running the steps in reverse to produce the original message, provided the same method is used to obtain the probability distribution. Much arithmetic as well as the decoding process have been left out for brevity. Readers wishing to fully understand the process are encouraged to seek out an example elsewhere. Because the lower bound of the subinterval is inclusive, we can simply use 0.687895 to represent our encoded message.\\

Unfortunately our example produces an output sequence that is longer than its input. This is not a fault of arithmetic coding, but a symptom of inaccurate probability estimates. As our model was arbitrary, it's expected that we would see poor output. For a more successful example of arithmetic coding, consider that the message ``AAAAA!'' can be coded as 0.0024. This represents a reduction of two symbols -- not counting the ever-present leading zero and the decimal.\\

Modeling is the process of generating a probability distribution estimate for the input sequence. Models can be static (as above) or dynamically generated. Dynamic models allow for continuous updates to probability values in response to the symbols observed in the sequence. A na{\"i}ve approach to dynamic modeling would be to initially consider all symbols equally probable, updating probabilities accordingly as symbols are processed.\\

Neural networks are dynamic models that update their probability distribution estimates based on dependencies learned from contexts. In the case of Mahoney's model, only spatially local contexts can be learned. By using a recurrent network architecture, spatial as well as temporal contexts can be used for dependency modeling. Utilizing more effective neural network architectures allows for the construction of more accurate language models.

\vspace{-0.4cm}
\subsection{Recurrent Neural Networks}
Recurrent neural networks are a class of neural networks well suited for modeling temporal systems such as sequences of audio or text. RNNs excel in these domains due to memory provided by recurrence in their hidden layers that allows them to learn dependencies over arbitrary time intervals. We can represent the hidden state of a RNN with a simple set of recurrence equations:
\begin{align*}
net_j(t) &= \sum_i^nx_i(t)v_{ji} + \sum_h^m y_h(t-1)u_{jh} + \theta_j\\
y_j(t) &= f\left(net_j(t)\right)
\end{align*}

where $y_j(t)$ is the output of the hidden state (layer) at time $t$ and $y_h(t-1)$ is the hidden state output from the previous time interval. The vectors $\mathbf{x}, \mathbf{V}, \mathbf{U}$ and $\mathbf{W}$ are the input, input-hidden, hidden-hidden, and hidden-output weights, respectively. Each layer is assigned an index variable (with notation borrowed from this guide~\cite{boden2002guide}) -- $k$ for output nodes, $j,h$ for hidden and $i$ for input nodes. The functions $f$ and $g$ (used later) are differentiable, nonlinear activation functions such as the sigmoid or hyperbolic tangent function. $\theta_j$ is a bias.\\
\pagebreak

The output state $y_k(t)$ can be computed as:

\begin{align*}
net_k(t) &= \sum_j^m y_j(t)w_{kj} + \theta_k\\
y_k(t) &= g\left(net_k(t)\right)\\
\end{align*}

All together, we see that a single forward pass through the network can be calculated with the following recurrence:

\begin{align*}
y_k(t) = g\left( \sum_j^m \left( f\left(  \sum_i^nx_i(t)v_{ji} + \sum_h^m y_h(t-1)u_{jh} + \theta_j  \right) \right) w_{kj} + \theta_k \right)\\
\end{align*}

\subsection{Backpropagation Through Time}
To allow for learning over arbitrary intervals, error values must be backpropageted through time. We use the cross entropy error function in our model defined as:

\begin{align*}
C = \frac{1}{2} \sum_p^n H(d,y)
\end{align*}

for the $p$th sample in the training set of length $n$ and the cross entropy function, $H(p,q)$:
\begin{align*}
H(p,q) = - \sum_x p(x) \log q(x)
\end{align*}

Together, our error function is:

\begin{align*}
C = \frac{1}{2} \sum_p^n \left( - \sum_k^m d_{pk} \log (y_{pk}) \right)\\
\end{align*}

for $d$, the desired output of $m$ output nodes. Weight updates are proportional to the negative cost gradient with respect to the weight that is being updated, scaled by the learning rate, $\eta$:
\vspace{-0.05cm}
\begin{align*}
\Delta w = - \eta \frac{\partial C}{\partial w}\\
\end{align*}

We can then compute the output error, $\delta_{pk}$ and hidden error $\delta_{pj}$, which can be backpropagated through time to obtain the error of the hidden layer at the previous time interval. \\

Indices $h$ and $j$ are for nodes sending and receiving the activation, respectively.

\begin{align*}
\delta_{pk} &= \frac{\partial C}{\partial y_{pk}} \frac{\partial y_{pk}}{\partial net_{pk}}\\
\delta_{pj} &= - \left( \sum_k^m \frac{\partial C}{\partial y_{pk}} \frac{\partial y_{pk}}{\partial net_{pk}} \frac{\partial net_{pk}}{y_{pj}}\right) \frac{\partial y_{pj}}{\partial net_{pj}}\\
& = \sum_k^m \delta_{pk}w_{kj}f'(y_{pj}) \\
\delta_{pj}(t-1) &= \sum_h^m \delta_{ph}(t)u_{hj}f' \left( y_{pj}(t-1) \right)\\
\end{align*}

\subsection{Gated Recurrent Units}
When backpropagating over many time intervals, error gradients tend to either vanish or explode. That is, the derivatives of the output at time $t$ with respect to unit activations at $t_0$ rapidly approach either zero or infinity as $t$ increases~\cite{bengio1993problem}. A popular solution to this problem is to use a Gated Recurrent Unit (GRU)~\cite{cho2014learning} -- a recurrent unit that adaptively resets its internal state. Networks of gated recurrent units allow for modeling dependencies at multiple time scales of arbitrary length, retaining both long and short-term memory. A single GRU consists of a hidden state along with reset and update gates.\\

When the reset gate, $r_j$ is closed ($r_j=0$), the value of the GRU's previous hidden state is ignored, effectively resetting the unit. The value of the reset gate is computed as:

\begin{align*}
r_j = \sigma \left( v_{jr} x_i + u_{jr} y_{h}(t-1)\right)\\
\end{align*}
\pagebreak

for the sigmoid activation function \mbox{$\sigma(t) = \left(1+e^t\right)^{-1}$}, the unit's input and previous hidden state, $x_i$ and $y_{h}(t-1)$, respectively. The weight matrices $\mathbf{V}$ and $\mathbf{U}$ follow from our previous equations.\\

The update gate $z_j$ is similar:

\begin{align*}
z_j = \sigma \left( v_{jz} x_i + u_{jz} y_{h}(t-1)\right)\\
\end{align*}

The new hidden state\footnote{Note the role of the reset gate in the calculation of the new hidden state.}, $\widetilde{y}_{j}(t)$ is:

\begin{align*}
\widetilde{y}_{j}(t) = \tanh \left( v_{j} x_{j} + u_j\left(r_j y_{h}(t-1)\right) \right) \\
\end{align*}

Finally, the unit's activation function, $y_{j}(t)$ can be calculated as a linear interpolation between the previous and current states:

\begin{align*}
y_{j}(t) = z_j y_{h}(t-1) + (1-z_j)\widetilde{y}_{j}(t) \\
\end{align*}

Cho et al. note that short-term dependencies are captured by units with frequently active reset gates, while long-term dependencies are best captured by units containing an active update gate.\\

The output of a single forward pass in a single layer GRU network can be represented using notation from the previous simple recurrent model:

\begin{align*}
net_k(t) &= \sum_j^m y_j(t)w_{kj} + \theta_k\\
  &= \sum_j^m \left( z_j y_{h}(t-1) + (1-z_j)\widetilde{y}_{j}(t)\right) w_{kj} + \theta_k\\
y_k(t) &= g\left(net_k(t)\right)\\
\end{align*}

\section{Model Architecture}
A close reader will notice that the topics covered in the background section address a succession of problems. We illustrated the need for an effective probabilistic model when compressing text data, then discussed the current state-of-the-art neural network architecture for generating such a model.\\

This section will address the issue of improving upon a vanilla GRU network architecture that operates solely on character sequences. The improvements discussed occur at a higher level of abstraction than the gate level architectures previously described, as we are seeking to build a practical model rather than propose a new recurrent unit architecture.\\
 
Table 3 outlines notation for the layers used in our architecture. Note that our model has two separate input layers. A graphical overview of our architecture can be found at the end of this section in Figure 1. 

\begin{table}[h]
\centering
\begin{tabular}{|c|l|}
\hline
Layer & Description                                                                    \\ \hline
$\mathbf{x}^{\langle c \rangle}(t)$     & Character input layer                        \\
$\mathbf{x}^{\langle p \rangle}(t)$     & POS input layer                              \\
$\mathbf{y}^{\langle c \rangle}(t) $    & GRU layer (character)       \\
$\mathbf{y}^{\langle p \rangle}(t)$     & GRU layer (POS)             \\
$\mathbf{y}^{\langle c|p \rangle}(t-1)$ & Previous hidden layer                        \\
$\bm{\Xi}^{\langle c|p \rangle}(t)$                            & Dropout layer                                \\
$\bm{\Psi}^{\langle c,p \rangle}(t)$                         & Merge layer                                  \\
$\mathbf{y}^{\langle \Psi \rangle}(t)$     & GRU layer (merged)          \\
$\mathbf{y}^{\langle D1 \rangle}(t)$    & Dense layer: RELU    \\
$\mathbf{y}^{\langle D1 \rangle}(t)$    & Dense layer: Softmax \\
$\mathbf{y}^{\langle out \rangle}(t)$                                  & Network output                               \\ \hline
\end{tabular}
\caption{Notations used in our RNN architecture.}
\end{table}

The character input layer, $\mathbf{x}^{\langle c \rangle}(t)$, is a \mbox{$40 \times 256$} one-hot representation of forty character sequences. This layer is paralleled by a second input layer containing part of speech information obtained from SyntaxNet. The part of speech tag (POS) input layer, $\mathbf{x}^{\langle p \rangle}(t)$ is a $40 \times 49$ one-hot\footnote{One-hot encoding is a way of representing information in which an array contains a single high bit (1) with the remaining bits low (0).} representation of part of speech tag sequences, each of which correspond to the character at the same respective index in the other input layer. \\

GRU layers $\mathbf{y}^{\langle c \rangle}(t)$ and $\mathbf{y}^{\langle p \rangle}(t)$ are also parallel. We will use the notation $\mathbf{y}^{\langle c|p \rangle}(t)$ when discussing separate but identical operations to both layers. Our implementation utilizes the hard (linearly approximated) sigmoid function in place of the standard logistic sigmoid as the GRU's inner activation function in order to reduce computational requirements. The outer activation, $g$ is the hyperbolic tangent function applied element-wise for each node in the layer. A forward pass through $\mathbf{y}^{\langle c \rangle}(t)$ and $\mathbf{y}^{\langle p \rangle}(t)$ is calculated as:

\begin{align*}
net^{\langle c|p \rangle}_j(t) &= \sum_i^n \left[ \left( z_i y_{h}(t-1) + (1-z_i)\widetilde{y}_{i}(t)\right) v_{ji} + \theta_j\right]^{\langle c|p \rangle}\\
y^{\langle c|p \rangle}_j(t) &= f\left(net^{\langle c|p \rangle}_j(t)\right)\\
\end{align*}

To prevent overfitting, dropout layers~\cite{srivastava2014dropout} $\bm{\Xi}^{\langle c \rangle}(t)$ and $\bm{\Xi}^{\langle p \rangle}(t)$ are applied to $\mathbf{y}^{\langle c \rangle}(t)$ and $\mathbf{y}^{\langle p \rangle}(t)$, respectively. The output of the dropout layers is a replica of the input, with the exception that output from a fractional number of nodes, randomly selected with probability $\rho$ is pinned to zero. After applying dropout, the state of the model is as follows:

\begin{align*}
\Xi^{\langle c|p \rangle}_j(t) &= \xi \left(y^{\langle c|p \rangle}_j(t)\right)\\
\text{where } \xi(x) &= \begin{cases} 
      0 & \text{ with probability } \rho\\
      x & \text{ otherwise}
   \end{cases}\\
\end{align*}
 
A merge layer, $\bm{\Psi}^{\langle c,p \rangle}(t)$ is applied to the output of the two dropout layers. This layer is a simple vector concatenation, represented here by the $||$ operator.
\enlargethispage{\baselineskip}
\begin{align*}
\bm{\Psi}^{\langle c,p \rangle}(t) = \bm{\Xi}^{\langle c \rangle}(t) \; || \; \bm{\Xi}^{\langle p \rangle}(t)\\
\end{align*}

The merged output feeds into a final GRU layer, $\mathbf{y}^{\langle m \rangle}(t)$ followed by two fully connected layers,  $\mathbf{y}^{\langle D1 \rangle}(t)$ and  $\mathbf{y}^{\langle D2 \rangle}(t)$ to produce the network output $\mathbf{y}^{\langle out \rangle}(t)$.

\begin{align*}
net^{\langle \Psi \rangle}_j(t) &= \sum_i^n \left[ \left( z_i y_{h}(t-1) + (1-z_i)\widetilde{y}_{i}(t)\right) w_{ji} + \theta_j\right]^{\langle \Psi \rangle}\\
y^{\langle \Psi \rangle}_j(t) &= f \left( net^{\langle \Psi \rangle}_j(t) \right) \\
\end{align*}

Fully connected (dense) layers are non-recurrent neural layers in which each node is connected to every node in both the preceding and following layer. Appending two fully connected layers to a recurrent neural networks was found to improve accuracy of speech models by transforming the sequential output of the recurrent layers to a more discriminatory space~\cite{sainath2015convolutional}. Adding two dense layers to our model had similar results, suggesting that the effect translates to sequence data from arbitrary domains.\\

The first dense layer, $\mathbf{y}^{\langle D1 \rangle}(t)$ uses the rectifier activation function $\text{ReLU}(x)$. This function is analogous to a half-wave reduction in digital signal processing, and has the advantage of being less computationally demanding than the sigmoid function. 
\begin{align*}
net^{\langle D1 \rangle}_j(t) &= \sum_j^m \left[y_j(t)w_{j} + \theta_j \right]^{\langle \Psi \rangle} \\
y^{\langle D1 \rangle}_j(t) &= \text{ReLU}\left( net^{\langle D1 \rangle}_j(t) \right) \\
\text{where ReLU}(x) &= \max(0,x)
\end{align*}

The second dense layer $\mathbf{y}^{\langle D2 \rangle}(t)$ employs a softmax activation that transforms the output of $\mathbf{y}^{\langle D1 \rangle}(t)$ from an arbitrary range to the interval [0,1] such that the sum of the 256 output nodes\footnote{There are 256 ASCII characters.} is 1. This is desirable as it allows our network output to satisfy the requirements of a proper probability mass function\footnote{ \mbox{$\sum_{x \in A} f_X(x)=1$}}. 
\Needspace{0.5\baselineskip}
\begin{align*}
net^{\langle D2 \rangle}_j(t) &= \sum_j^m \left[y_j(t)w_{j} + \theta_j \right]^{\langle D1 \rangle} \\
y^{\langle D2 \rangle}_j(t) &= \text{softmax}\left( net^{\langle D2 \rangle}_j(t) \right) \\
\text{softmax}(x) &= e^x \left(\sum_{n}^m e^{x_n}\right)^{-1}
\end{align*}

The network output, $y^{\langle out \rangle}_k(t)$ is simply the output of the final dense layer, $y^{\langle D2 \rangle}_k(t)$.

\begin{align*}
y^{\langle out \rangle}_k(t) &= y^{\langle D2 \rangle}_j(t)  
\end{align*}

To keep calculation simple, we've been operating on individual neural units. As we've reached the output layer, it's important to remember that we're working with vectors:

\begin{align*}
\mathbf{y}^{\langle out \rangle}(t) &= \left[y^{\langle out \rangle}_0(t), ..., y^{\langle out \rangle}_k(t)\right] \\
\end{align*}

We now see why the softmax activation function is critical to the model -- the network's output always sums to one and is a valid representation of probability estimates for each character:
\begin{align*}
\sum_{i=0}^j \left[ y^{\langle out \rangle}_i(t), ..., y^{\langle out \rangle}_k(t) \right] &= 1\\
\end{align*}

Illustrating a full forward pass through this network would provide little value to the reader and require a significant amount of space. By following the layer descriptions in this section, we've essentially already completed the forward pass. \\

Backpropagation for an architecture of this complexity is not an easy task. Fortunately, automatic differentiation frees us from the burden of calculating the error gradient. Our implementation utilized Keras~\cite{Keras}, a wrapper for Theano~\cite{bergstra2010theano}. Readers seeking information on the gradient calculations should consult the Theano documentation. \\

\begin{figure*}[!h]
\centering
\makebox[0pt]{\includegraphics[scale=0.4]{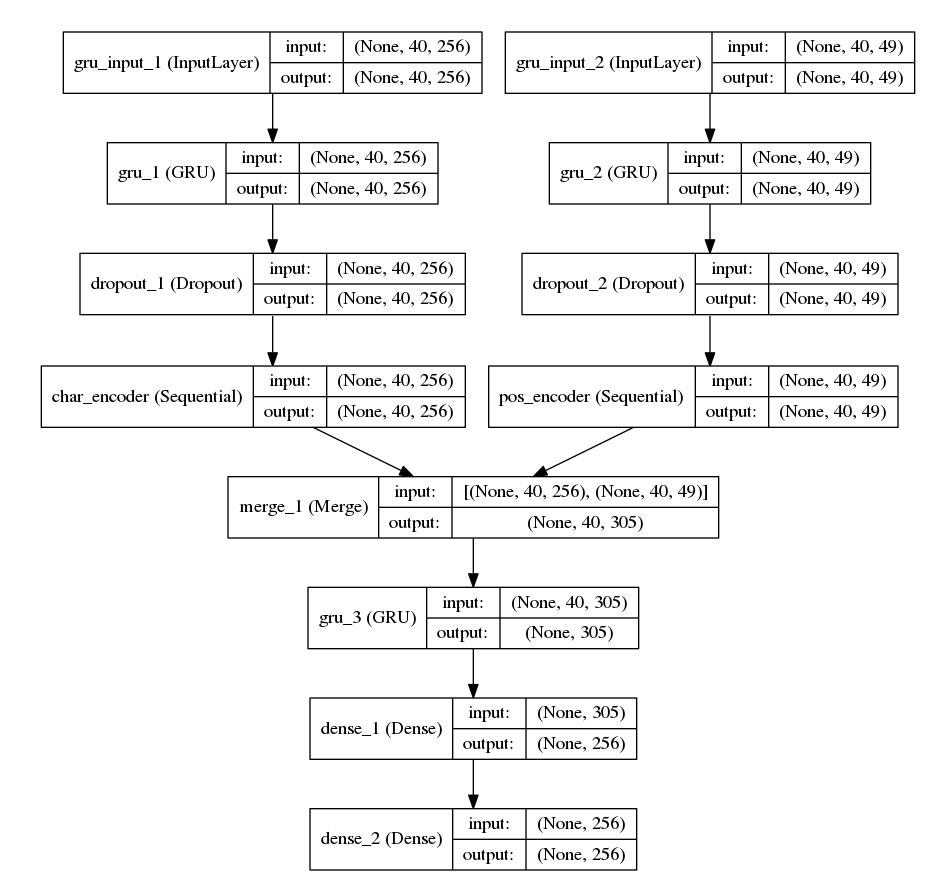}}
\caption{Architectural overview.}
\end{figure*}

\section{Training and Evaluation}

Training data was obtained from Project Gutenberg~\cite{hart1971project}. Models were trained on single books  -- preserving the single stream, single model training method used by Mahoney. The input text was passed through SyntaxNet to obtain part of speech tags information for each word. Additionally, input was split into 40 character chunks with a sliding window. Part of speech tags were replicated such that each character in a word was given the appropriate tag for the word. The 41st character in each window was used as the target output. It's also worth noting that this system is based loosely upon the \texttt{lstm\_text\_generation} example from the Keras library. Readers seeking to build upon our work should consult this example. \\

RMSprop~\cite{tieleman2012lecture} was used to optimize gradient descent. RMSprop keeps a moving average of the gradient squared for each weight as shown:

\begin{align*}
E\left[\left( \frac{\partial C}{\partial w} \right) ^2\right]^{\langle t \rangle}\!\!\!\!\!\!\! &= 0.9E\left[\left( \frac{\partial C}{\partial w} \right) ^2\right]^{\langle t-1 \rangle}\!\!\!\!\!\!\!\!\!\!\!\! + 0.1\left[\left( \frac{\partial C}{\partial w} \right) ^2\right]^{\langle t \rangle}\\
\theta^{\langle t + 1\rangle} &= \theta^{\langle t \rangle} - \frac{\eta}{\sqrt{E\left[\left( \frac{\partial C}{\partial w} \right) ^2\right]^{\langle t \rangle}} + \epsilon}\left[\frac{\partial C}{\partial w}\right]^{\langle t \rangle}\\
\end{align*}

Models were trained on four books of various length for a minimum of 700 epochs per document. Variation in the number of training iterations was due to the increased computation required to model longer  documents. For comparison with the referenced LSTM mode, we also trained a model on \textit{The complete works of Friedrich Nietzsche}. The length of all documents used in training is illustrated in Table 4.\\

\begin{table}[h]
\centering
\caption{Training document length.}
\label{my-label}
\begin{tabular}{|l|c|}
\hline
Document                  & Length (characters) \\ \hline
alice.txt                 & 167518              \\
holmes.txt                & 581878              \\
netzsche.txt              & 600901              \\
pride\_and\_prejudice.txt & 704145              \\
two\_cities.txt           & 776646              \\ \hline
\end{tabular}
\end{table}

To quantify the effect of part of speech information, models for \textit{Pride and Prejudice} were trained with and without part of speech tags\footnote{To accomplish this, we simply set the part of speech input vector to zero.}.  \\

Amazon \texttt{g2.2xlarge} EC2 instances were used to perform training and evaluation. For longer documents, each epoch took approximately 230 seconds, equating to roughly 48 hours of computation per document. We've provided a preconfigured AMI for those wishing to verify or expand\footnote{A full copy of our codebase is available at \texttt{https://github.com/davidcox143/rnn-text-compress}} upon our results without going through the trouble of resolving software dependencies. The AMI is publicly available as \texttt{ami-2c3a7a4c}. \\



\section{Results}

All models converged to a high level of accuracy within the training window. Figure 2 illustrates convergence and raises some noteworthy discussion points\footnote{Figures 2 and 3 are located in the Appendix.}. Unsurprisingly, the shortest document in our training set converged in the fewest number of epochs and attained the highest level of accuracy. This near-perfect accuracy is indicative of severe overfitting, and implies that our model is capable of essentially memorizing documents less than $167\!,500$ characters in length. Overfitting would be undesirable if training a general language model, but poses less of a concern in our usage case. \\

The model trained on \textit{A Tale of Two Cities} exhibits gradient instability after epoch 650, significantly reducing its accuracy from that point onwards. Unstable gradients can occur when converged models are allowed to continue training, as appears in this case. This example highlights the sometimes chaotic~\cite{beer1995dynamics} behavior of recurrent neural networks. Gated recurrent units often produce relatively stable models; however, their dynamics remain poorly understood. An in-depth analysis of GRU network dynamics would likely shed light on the observed long-term instability. \\

The addition of part of speech information to the \textit{Pride and Prejudice} model resulted in an average accuracy increase of $5.33 \%$, as shown in Figure 3.\\

Further exploration of this metric was not performed due to computational and time constraints on the project. As consolidation, we considered the generalization performance of our document-specific models and found them to be reasonably accurate when applied to the other training documents. The generalization performance of the \textit{Pride and Prejudice} model is shown in Table 5.

\begin{table}[h]
\centering
\caption{Model generalization performance.}
\label{my-label}
\begin{tabular}{|l|c|}
\hline
Document                  & Accuracy (\%) \\ \hline
alice.txt                 & 35.99         \\
holmes.txt                & 59.04         \\
netzsche.txt              & 58.44         \\
pride\_and\_prejudice.txt & 93.91         \\
two\_cities.txt           & 54.42         \\ \hline
\end{tabular}
\end{table}


\section{Discussion}

The results of this effort make a strong case for a pre-trained, generalized language model that could be used in text compression. Document-specific compression benchmarks were not performed, as such metrics are slightly outside the scope of this publication. Proper compression benchmarks for a generalized model will be the focus of future work. \\

We anticipate that model performance could be further increased by utilizing word dependency trees provided by SyntaxNet. The combination of semantic and syntactic information would likely allow for the representation of more complex word relationships than syntactic information alone. While accuracy does not seem to be of concern for single stream, single model usage contexts such as ours, generalized models stand to benefit from the understanding of complex contextual relationships derived from semantic information.  \\ 

The computational overhead associated with training a model for even as few as 100 epochs limits the practicality of our current implementation. Use of a general, pre-trained model would  eliminate this problem -- the time required to compute a single forward pass for the prediction of the next character is negligible. \\

Training a generalized model will require significantly more computational resources. Generalized models require a large number of diverse training documents. The one-hot encoding used in our architecture is not memory efficient by design. Even if batch training is used to alleviate memory requirements, training time would far exceed the 48 hours required to train a document-specific model.\\  

This work also raises the interesting concept of utilizing the output of one neural network as the input to another. \\

The composition of neural networks can be performed in a fashion similar to the composition of functions. This should be almost intuitive, as the forward pass through a neural network is in fact a function. Neural network composition may prove to be a critical area of machine learning research. Using separately trained, domain-specific neural networks is likely a better approach to complex tasks such as language modeling than training a single, monolithic network.

\nocite{*}
\bibliographystyle{acm}
\bibliography{rnn_text_compress}

\onecolumn
\section*{Appendix}
\centering
\makebox[0pt]{\includegraphics[scale=1.1]{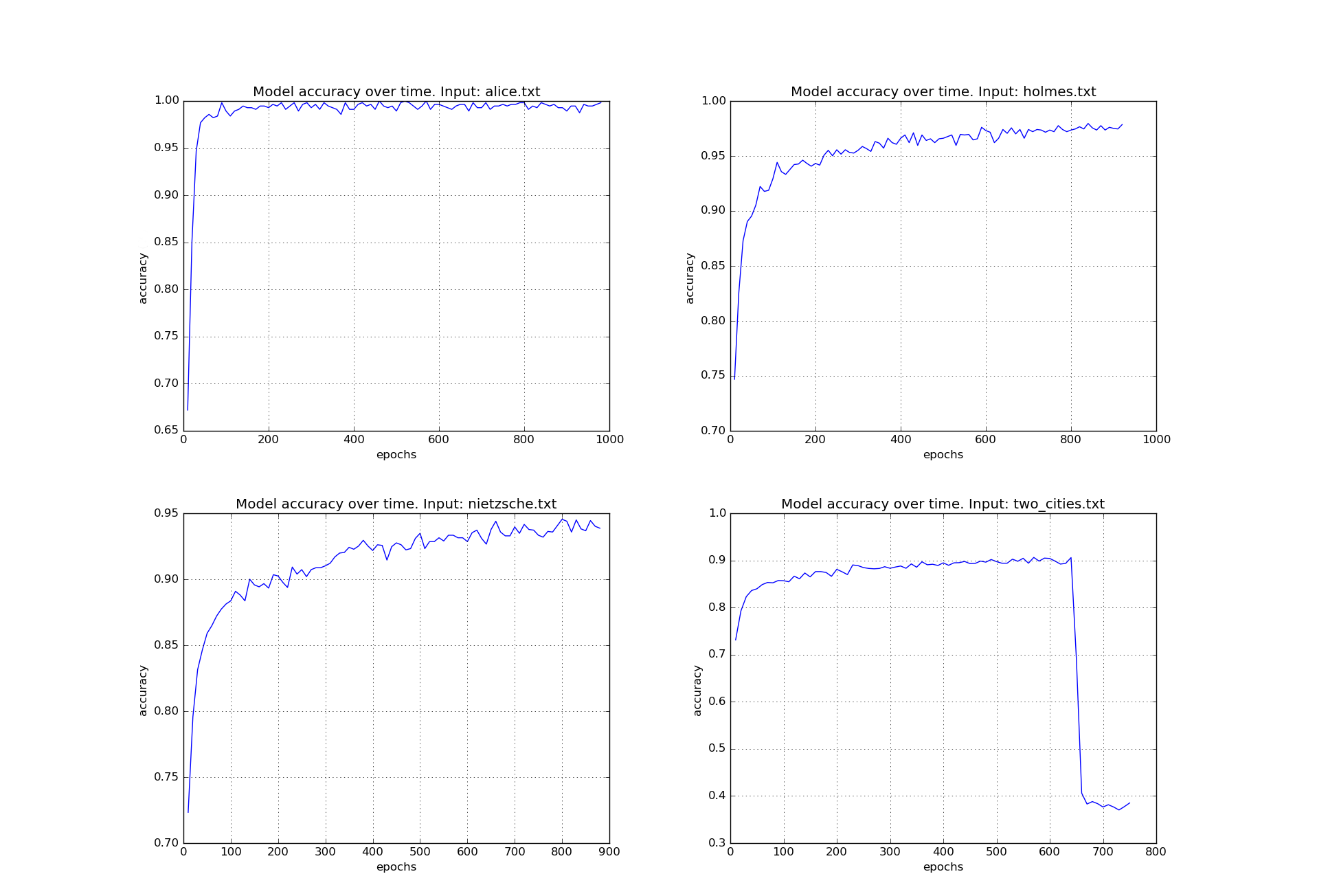}}
\captionof{figure}{Model performance over time.}
\vspace{1cm}
\makebox[0pt]{\includegraphics[scale=0.55]{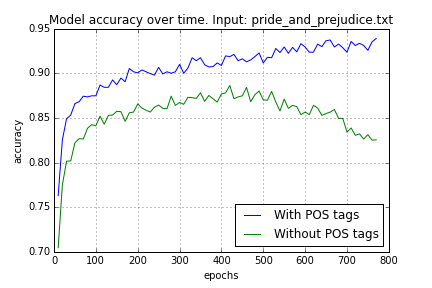}}
\captionof{figure}{Impact of part of speech information on model accuracy.}


\end{document}